\documentclass[letterpaper, 10 pt, conference]{ieeeconf}  % Comment this line out if you need a4paper

\IEEEoverridecommandlockouts                              % This command is only needed if 
\usepackage{graphicx} % for pdf, bitmapped graphics files
\usepackage{amsmath} % assumes amsmath package installed
\usepackage{amssymb}  % assumes amsmath package installed
\usepackage{subcaption}
\usepackage{siunitx}
\usepackage{cite}
\usepackage{bm}
\usepackage{mathtools}
\usepackage[font=footnotesize]{caption}
\usepackage[abbreviations]{glossaries-extra}
\usepackage{flushend}
\usepackage{booktabs}
\usepackage[colorlinks = true,
            linkcolor = blue,
            urlcolor  = blue,
            citecolor = blue,
            anchorcolor = blue]{hyperref}
\usepackage[capitalise]{cleveref}
\usepackage{algorithm}
\usepackage{algpseudocode}
\usepackage{multirow}
\usepackage{makecell}
\usepackage{gensymb}

\title{\LARGE \bf ORB-SfMLearner: ORB-Guided Self-supervised Visual Odometry with Selective Online Adaptation}

\author{Yanlin Jin$^{1}$, Rui-Yang Ju$^{2}$, Haojun Liu$^{3}$, Yuzhong Zhong$^{4,*}$% <-this % stops a space
\thanks{$^{1}$Yanlin Jin was with the College of Electrical
Engineering, Sichuan University and is now with Rice University. {\tt\footnotesize neil.yl.jin@gmail.com}}%
\thanks{$^{2}$Rui-Yang Ju with the Graduate Institute of Networking and Multimedia, National Taiwan University.   {\tt\footnotesize jryjry1094791442@gmail.com}}%
\thanks{$^{3}$Haojun Liu with the Language Technologies Institute, Carnegie Mellon University. {\tt\footnotesize haojunli@andrew.cmu.edu}}%
\thanks{$^{4}$Yuzhong Zhong with College of Electrical Engineering, Sichuan University. $^{*}$Corresponding author: {\tt\footnotesize zyzc122@163.com}}%
\thanks{This work has been accepted to ICRA 2025 for publication. Copyright may be transferred without notice, after which this version may no longer be accessible.}
}

\begin{document}

% This allows you to play around with how the references show up in the paper -- can be used to make space :))
\bstctlcite{IEEEexample:BSTcontrol}

\maketitle
\thispagestyle{empty}
\pagestyle{empty}

%%%%%%%%%%%%%%%%%%%%%%%%%%%%%%%%%%%%%%%%%%%%%%%%%%%%%%%%%%%%%%%%%%%%%%%%%%%%%%%%
\begin{abstract}
Deep visual odometry, despite extensive research, still faces limitations in accuracy and generalizability that prevent its broader application. To address these challenges, we propose an Oriented FAST and Rotated BRIEF (ORB)-guided visual odometry with selective online adaptation named ORB-SfMLearner. We present a novel use of ORB features for learning-based ego-motion estimation, leading to more robust and accurate results. We also introduce the cross-attention mechanism to enhance the explainability of PoseNet and have revealed that driving direction of the vehicle can be explained through the attention weights. To improve generalizability, our selective online adaptation allows the network to rapidly and selectively adjust to the optimal parameters across different domains. Experimental results on KITTI and vKITTI datasets show that our method outperforms previous state-of-the-art deep visual odometry methods in terms of ego-motion accuracy and generalizability.
Code is available at
\begingroup
\hypersetup{urlcolor=magenta}
\normalfont
\url{https://github.com/PeaceNeil/ORB-SfMLearner}.
\endgroup
\end{abstract}

\section{INTRODUCTION\label{sec:intro}}
Estimating camera ego-motion from monocular videos is essential for various computer vision and robotics tasks. 
Recent advances in 3D representation learning~\cite{mildenhall2021nerf, kerbl3Dgaussians} have also heightened the demand for camera pose estimation. 
Traditional methods~\cite{schoenberger2016sfm, mur2015orb, mur2017orb, campos2021orb} find matches across frames and restore camera transformations with epipolar geometry, while learning-based self-supervised methods~\cite{zhou2017unsupervised, jia2021scsfm, godard2019digging} usually infer depth and ego-motion simultaneously and then establish a self-supervised constraint with the photometric reconstruction error. 
Learning-based methods have been widely studied in recent years due to its fast inference and ability to learn high-level features from data~\cite{li2019sequential, yin2018geonet}.

% rewrite para 2 and 3
However, learning-based visual odometry (VO) still faces several challenges. 
First, the accuracy of depth and ego-motion estimation remains inferior to that of traditional methods. 
Second, due to the black-box nature of neural networks, the decision-making process is not well understood, which reduces trust in the system. 
Finally, and most importantly, learning-based VO suffers great performance drop on unseen test scene because of the common large domain gap. 
Even within the domain of autonomous driving, factors such as vehicle speed and weather changes can have a significant impact. 
We even find that the model's performance is poor on some already seen training samples that display a certain domain gap, indicating its limited generalization capability and also fitting ability. 

To address these problems, several works have been proposed.
Training on larger datasets may help mitigate effect of domain gaps and achieve better accuracy. 
Wang~\emph{et al.}~\cite{pmlr-v155-tartan} use synthetic data to attempt large-scale VO training. 
However, due to the complexity of real-world environments, it's challenging to gather a sufficient amount of data, which also requires a significantly longer training time.
The self-supervised nature of current learning-based methods provides another solution. 
Some recent works~\cite{vodisch2023covio, li2020self} focus on online fine-tuning of pre-trained VO models during test time. This learning while testing approach proves to be very effective.
% Therefore, we also propose a new online adaptation algorithm based on this principle. 
However, in scenarios where directly training on the entire test set fails to yield satisfactory results, online fine-tuning hardly works well. 
Therefore, more robust training strategies are still needed for the model to achieve better performance independently.

This work demonstrates several simple yet effective approaches to develop a more generalizable and explainable deep VO estimation system. 
% add insights here, why orb, therefore guidance
We notice that the input images may vary in style due to factors like lighting and weather. 
Therefore, we aim to guide the network's attention to more stable features.
Inspired by traditional Simultaneous Localization and Mapping (SLAM) methods ORB-SLAM~\cite{mur2015orb, mur2017orb, campos2021orb}, we incorporate the pipeline with ORB~\cite{rublee2011orb} features augmentation. %, that overcome domain difference caused by appearance change of image data and also perform as guidance for network's attention. 
We further explored the influence of ORB features by designing cross-attention layers within PoseNet, and the results provide a compelling explanation of ORB guidance. 
Building on our ORB-guided VO, we further propose selective online adaptation to enhance its generalizability.
We demonstrate the effectiveness of our methods with ablation studies and our evaluation results outperform previous monocular self-supervised state-of-the-art (SOTA) VO works.% in terms of both camera poses estimation and depth estimation.
\begin{figure*}[t]
    \centering
    \includegraphics[width=1\linewidth]{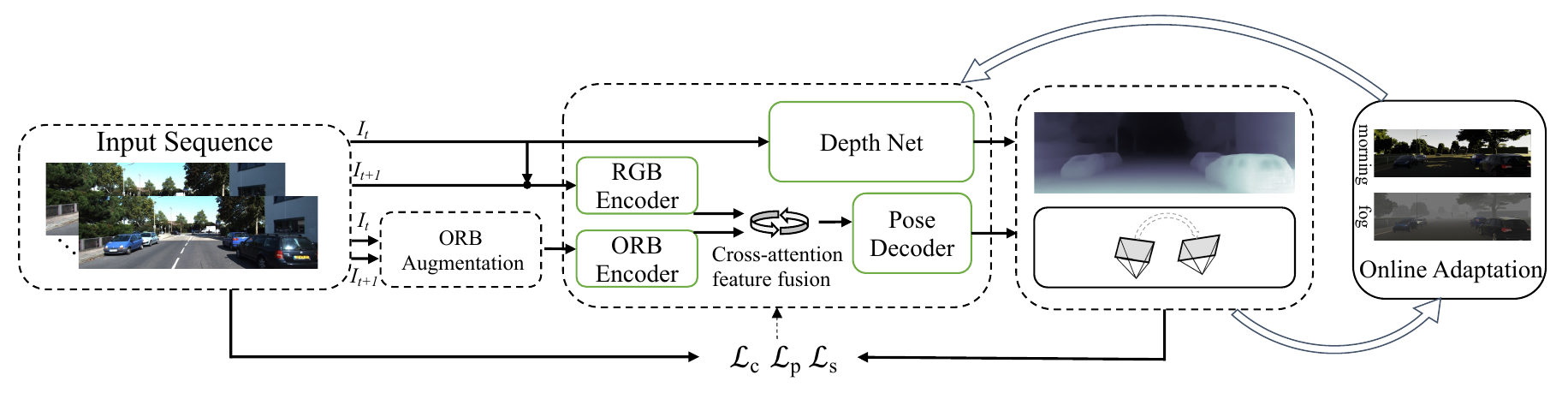}
    \caption{The pipeline of ORB-SfMLearner. The VO DepthNet estimates depth, while the PoseNet estimates the relative pose between two frames after fusing ORB and RGB features through a multi-head cross attention mechanism. 
    The network is trained using the self-supervised reprojection error \(\mathcal{L}_p\)~\cite{zhou2017unsupervised}, geometry consistency error \(\mathcal{L}_c\)~\cite{bian2019unsupervised}, and depth map smoothness error \(\mathcal{L}_s\)~\cite{godard2019digging}. 
    During inference, the network selectively performs online adaptation, learning to use the weights most suitable for the current scene, thereby achieving good generalization in challenging conditions, such as foggy weather.}
    \label{fig:pipeline}
\end{figure*}
To summarize, our contributions are:
\begin{itemize}
\item{We propose an effective and simple ORB augmentation method for self-supervised VO learning that boosts its accuracy. 
Our PoseNet learns from ORB features and achieves SOTA ego-motion estimation on the KITTI dataset. 
This neat augmentation method shows its potential to be applied in a broader range of vision tasks.} 
\item{To enhance interpretability of the networks' learning process, we intuitively explored the impact of ORB features. 
As one of the earliest works to explore the interpretability of ego-motion estimation, we aim to provide insights for related research.} 
\item{We optimize our VO system for generalizability during both training and online adaptation phases. 
Our online adaptation strategy enables rapid optimization of network parameters based on current data and selects the optimal parameters to output refined estimation.} 
\end{itemize}

% Paper Organization

\section{RELATED WORKS}
\textbf{Visual Odometry} is a technique for robots to locate themselves with image stream from visual sensors. 
As a crucial part in the SLAM system, the principal task for VO is to provide camera pose estimation, also known as the ego motion estimation. In the context of feature-based SLAM~\cite{engel2014lsd,mur2015orb,mur2017orb,campos2021orb}, previous work such as ORB-SLAM~\cite{mur2015orb} chooses to use ORB features, which offer robust pose recognition ability and efficient real-time tracking. 
Similar to classical direct VO~\cite{engel2014lsd}, the core of recent learning-based self-supervised VO is to minimize photometric loss. 
SfMLearner~\cite{zhou2017unsupervised} was the first to jointly train a PoseNet and a DepthNet and optimize them based on photometric error. 
Monodepth2~\cite{godard2019digging} continues this approach but uses ResNet and U-Net architectures for feature extraction and decoding outputs for depth maps and poses. 
Additionally, Monodepth2 proposes an auto-mask strategy to ignore pixels that remain stationary between frames, which helps reduce the impact caused by synchronized camera and object movement, camera stillness, or weak object textures. 
SC-SfMLearner~\cite{jia2021scsfm} further introduces a geometric consistency loss, which calculates reprojection error by transforming the depth maps of adjacent frames based on the estimated pose. 
This constrains the continuity and structure of the depth maps across frames. 
These works are key baselines for subsequent self-supervised depth and pose estimation research. 
Building on the core supervision being the photometric error from reprojection, they contribute to areas such as depth map scale consistency and moving object masking, achieving satisfactory results on the KITTI~\cite{Geiger2012CVPR} benchmark.

\textbf{Explainability in Deep VO} has also been explored by some recent works, but mostly focusing on the rationale behind depth estimation. 
One commonly used method is feature visualization, such as CAM~\cite{cam} and Grad-CAM~\cite{selvaraju2016gradcam}, to highlight important regions in an image. 
In addition, Tom~\emph{et al.}~\cite{dijk2019neural} treat DepthNet as a black box and evaluate depth output in response to different input variations.
Specifically, they tested how factors like object position, occlusion, and types affect the model's predictions. 
Interestingly, removing the center portion of a car does not significantly impact the results, but if the edges of the car's bottom are also removed, the DepthNet might fail to recognize the car. 
For the pose estimation, Sattler~\emph{et al.}~\cite{sattler2019understanding} revealed that supervised absolute pose regression is essentially more related to image retrieval instead of geometry-based learning. 
Pose regression methods regress the camera pose of an input image, while self-supervised VO usually takes two consecutive frames and outputs a relative pose estimation. 
To the best of our knowledge, what the self-supervised PoseNet focuses on remains unexplored.

\textbf{Generalizability} is a crucial for making learning-based VO applicable in real-world environments. 
Li \emph{et al.}~\cite{li2020self} use LSTM~\cite{hochreiter1997long} to extract temporal and spatial information from the input data and enables the network to continuously optimize its parameters based on past experiences. 
CoVIO~\cite{vodisch2023covio} continuously updates network weights during inference because of its self-supervised nature, which is similar to our method. 
They designed a replay buffer, where new frames with cosine similarity to frames in the buffer below a certain threshold are added to the buffer. 
In contrast, we selectively update parameters based on the self-supervised loss.

\section{METHODS}
\subsection{Pipeline Overview}
We aim to build a VO pipeline that takes advantage of more stable features and adapt itself to overcome variance of testing scenarios. 
As shown in Fig.~\ref{fig:pipeline}, with two consecutive frames \((I_t, I_{t+1})\) from a monocular video as input, our pipeline first augments image data by extracting the ORB features. 
Then we input the original RGB data and the extracted ORB features into two separate encoders. We apply cross-attention to weight the importance of encoded ORB features relative to RGB features. 
The fused features are then fed into a decoder to predict the relative 6D camera pose between the two frames. % We then concatenate ORB features to original RGB data and input them to a pose net to predict the relative 6D camera pose between the two frames. 
Meanwhile, our DepthNet takes one original image input from \(I_t\) and \(I_{t+1}\) each time, and outputs the disparity estimation of the current frame. 

Based on the network output depth and the predicted relative pose transformation, we can project one frame to another. 
For instance, we synthesize the \(I_{t+1}'\) by projecting \(I_t\). 
Comparing it with the real \(I_{t+1}\), we compute a photometric reconstruction error to form the self-supervised constraint. 
This is the fundamental principle for self-supervised VO. 
Furthermore, in order to enforce depth scale consistency, we adopt the geometry consistency loss proposed in~\cite{jia2021scsfm}.

Finally, in the test phase, both the PoseNet and the DepthNet optimize their parameters based on the given input. 
Our online adaptation algorithm rapidly updates and chooses the most suitable parameters for estimating camera ego-motion and depth. 
Unlike during training, where we avoid overfitting, we actually aim for the model to overfit on the current tested snippet.

\subsection{Self-supervision Principle for VO\label{loss}}
The essence of self-supervised VO lies in reconstructing adjacent frames using the depth and inter-frame relative pose output by the neural network. 
The similarity between the real reconstruction target and the synthesized one reflects the quality of the estimated depth and pose. 
Given two frames \(I_a\) and \(I_b\), the depths \(D_a\) and \(D_b\) of them are predicted by DepthNet and their relative transformation \(T_{ab}\) by the PoseNet. 
We synthesize the reconstructed \(I_b'\) with the differentiable warping process proposed by Zhou~\emph{et al.}~\cite{zhou2017unsupervised} and choose \(\mathcal{L}_1\) and structural similarity (SSIM) loss to construct the photometric loss function \(\mathcal{L}_p\):
\begin{equation}
\begin{aligned}
\mathcal{L}_p\left(I\right) = \frac{1}{n} \sum_{i=0}^{n} \Bigg(&\lambda \left\|I_b(i) - I_b^{\prime}(i)\right\|_1 \\
&+ (1 - \lambda) \frac{1 - \operatorname{SSIM}\left(I_b, I_b^{\prime}\right)\left(i\right)}{2}\Bigg),
\end{aligned}
\label{eq:tnet_self}
\end{equation}
where \(I_b(i)\) and \(I_b^{\prime}(i)\) denote the pixel values at pixel \(i\) in the two images, and \(n\) represents the total number of pixels.
$\operatorname{SSIM}(I_b, I_b^{\prime})$ is the element-wise similarity map, subtracting it from 1 and dividing by 2 scales its range to $[0, 1]$.
Weight \(\lambda\) is set to 0.15 as in~\cite{godard2017unsupervised, yin2018geonet, bian2019unsupervised}.

In addition to the main photometric reconstruction constraint, we enforce depth consistency with the geometric consistency loss \(\mathcal{L}_c\) following~\cite{jia2021scsfm}. 
Similar to warping RGB frames, the depth map \(D_a\) is warped to \(D_b\) with the predicted transformation \(T_{ab}\). The depth inconsistency is as follows:
\begin{equation}
\begin{aligned}
&  \mathcal{L}_{c} = \frac{1}{n}\sum_{i=0}^{n}\frac{|{D}_{b}'(i)-{D}_{b}''(i)|}{{D}_{b}'(i)+{D}_{b}''(i)},
\label{eq:frames}
\end{aligned}
\end{equation}
where \(D_b'\) is the warped depth from \(D_a\), and \(D_b''\) is the interpolated \(D_b\). 
The geometric consistency constraint enforces the inter-frame depth continuity which eventually leads to the depth and ego-motion scale consistency in entire sequence. Additionally, a smooth constraint \(\mathcal{L}_s\)~\cite{godard2019digging} is applied to regularize the estimated
depth map.

\subsection{ORB Features Augmentation\label{ORB Aug}}
\begin{figure}[t]
    \centering
    \includegraphics[width=1\linewidth]{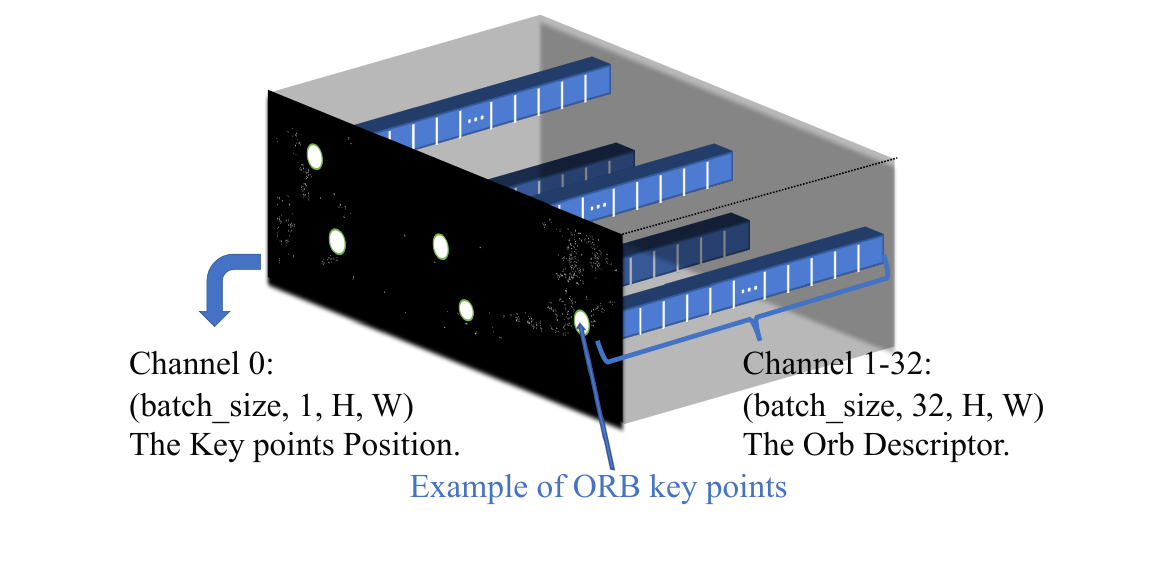}
    \caption{Our augmented ORB data structure.
    For each original RGB image, we extract and form its ORB features as a 33-channel tensor. 
    The first channel has the same dimension as original pictures but only feature points have value 1 to indicate key points positions. 
    The other 32 channels store the ORB descriptors behind the key points. %Only the image coordinate positions of the ORB key points exist descriptor values. 
    When in use, the two blocks of ORB tensors will be concatenated along the channel dimension.
    This method enables the representation of key points' positional information and potential matching relationships between detected key points in two images.}
    \label{fig:ORB}
    \vspace{-10pt}
\end{figure}

We use ORB features to efficiently augment image data for the PoseNet and guide the network's attention regions. 
ORB feature detection, compared to Scale-Invariant Feature Transform (SIFT)~\cite{lowe1999sift} and Speeded Up Robust Features (SURF)~\cite{bay2008speeded}, requires less computation. 
It consists of the Features from Accelerated Segment Test (FAST)~\cite{rosten2006machine} corner detector to determine key points position and the Binary Robust Independent Elementary Features (BRIEF)~\cite{calonder2010brief} descriptors to describe the key points features. 
For resized data from the KITTI dataset with dimensions of $832 \times 256$ pixels, we extract multi-scale oriented FAST corners with a scale factor of 2, using Harris score~\cite{harris1988combined} to choose the top 1,000 key points. 
We store each 256-bit binary descriptor in 32 bytes and then perform our ORB features augmentation.

We organize the ORB features into a tensor of 33 channels, with width and height matching the size of the input image, as shown in Fig.~\ref{fig:ORB}. 
The first channel contains only the positional information of detected ORB key points, where key points are represented by a value of 1, while non-key points are represented by 0. 
The rest channels contain the ORB descriptor along the channel axis, where each descriptor is stored behind the first channel's key points. 
The remaining elements are left as zeros. 

During training, we first try to concatenate each ORB block to its original 3-channel RGB input, and then construct a 72-channel augmented input by combining the two 36-channel blocks. 
Although this appears to be a simple concatenation, previous works in the field of content generation~\cite{saharia2022image, Brooks_2023_CVPR, li2023gligen} have shown that additional input in extra channels can effectively guide and ground the output. 
Similarly, this augmentation first guides network's attention by assigning key points areas larger weight values, which enables the network to focus more on stable features of input data. 
Then, observing along the channel axis, the deviations between two sets of 36-channel ORB features contain information about relative pose changes and the values of ORB descriptors describe the matching relationships between feature points. 
Those factors lead to our model's robustness in face of various domains of data.

\subsection{Explain the ORB Guidance}
To confirm that the ORB feature has indeed been learned by the network, we redesign a PoseNet embedded with multihead cross attention layers in an attempt to open this black box. 
Following~\cite{jia2021scsfm}, we design the new PoseNet based on ResNet-18. 
Instead of directly inputting the concatenated RGB and ORB together into the ResNet, we now use two separate ResNet encoders, one for RGB and the other one for ORB. %Fig.~\ref{} shows the structure of the proposed ORB-attention PoseNet. 
The cross-attention module takes RGB feature maps as key ($K$) and value ($V$), and ORB feature maps as query ($Q$). 
We project each feature map from 512 channels to 128 embedding dimensions, then rearrange them into sequence format and fed into the attention layer. 
With Equation~(\ref{eq:attn_weights}) and~(\ref{eq:attn_out}), we compute attention weights by performing scaled dot-product attention across the $n=8$ heads, followed by a softmax operation to normalize these weights.
The resulting weighted sum of the $V$ from each head is finally projected out to the original feature map dimension and concatenated with original RGB features. 
We modify the first convolution layer of the PoseNet's decoder so that the decoder can accept the concatenated features as input.
Our attention weights (Attn\_Weights) and final output are calculated as follows:
\begin{equation}
\text{Attn\_Weights}_{b,n,i,j} = \text{softmax}\left(\frac{Q_{b,n,i,d} \cdot K_{b,n,j,d}^T}{\sqrt{d_k}}\right),
\label{eq:attn_weights}
\end{equation}

\begin{equation}
\text{Output}_{b,n,i,d} = \sum_{j} \text{Attn\_Weights}_{b,n,i,j} \cdot V_{b,n,j,d}.
\label{eq:attn_out}
\end{equation}

\begin{figure}[t]
    \centering
    \includegraphics[width=1\linewidth]{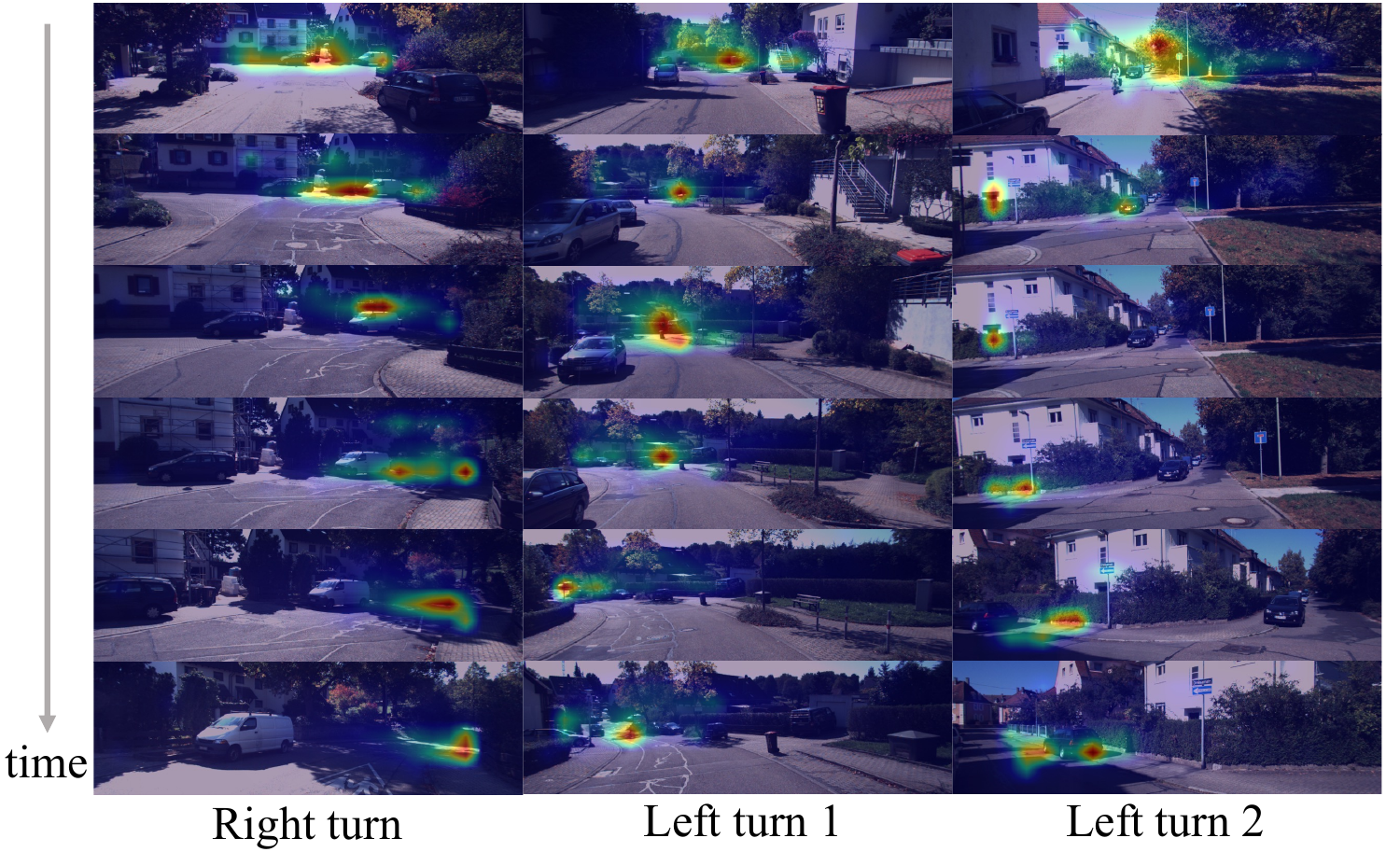}
    \caption{By visualizing the attention weights, a clear pattern emerges: during left or right turns, the regions with high weights also shift accordingly, often pointing towards the end of the road. 
    The two columns on the left are selected from KITTI Odometry sequence 09, and the rightmost column is selected from sequence 07.}
    \label{fig:cross_vis}
\end{figure}

The Cross Attention process we introduced integrates the RGB and ORB features. 
Its actual purpose is to use the ORB features to guide the network in focusing on the content in the RGB features. 
This is similar to the previous method where we concatenated RGB and ORB images before inputting them into the network. 
However, by visualizing the attention weights, we can now more intuitively demonstrate the network's preferences as shown in Fig.~\ref{fig:cross_vis}.

% \subsection{Dynammic object masking}

\subsection{Selective Online Adaptation\label{SOA}}
After pre-training the PoseNet and DepthNet on the source domain, we continue to adapt the model during inference with our selective online adaptation (SOA), as illustrated in Algorithm~\ref{alg:online}. 
We first prepare the incoming frames with the proposed ORB-Augmentation. 
Then we conduct inference with current PoseNet and DepthNet and use the estimated relative pose $P$ and depth $D_i, D_{i+1}$ to perform differentiable warping (Section \ref{loss}). 
Here, the loss from this self-supervised process is used to update the model parameters. 
However, we don't directly iterate to optimize and output the results. 
Instead, we infer with the updated model to compute the loss again, and only if the loss decreases do we select the updated parameters. 
We call this strategy selective adaptation.
For simplicity, Algorithm~\ref{alg:online} only shows the case where each input snippet consists of 2 frames. 
We can also input more frames and perform differentiable warping on each pair of frames, updating the model online using the average loss.

\begin{algorithm}
\caption{Selective Online Adaptation}
\label{alg:online}
\begin{algorithmic}[1]
\Require intrinsics $K$, Pre-trained $PoseNet$, $DepthNet$
\For{each consecutive pair $(I_i, I_{i+1})$ in test set}
    % \State back up $current\_params$
    \State Pre-process with ORB-Augmentation
    \For{$n = 0$ to $k$}
        \State Obtain $D_i$ and $D_{i+1}$ from $DepthNet$, and $P$ from $PoseNet$($I_i,I_{i+1}$)
        \State $error_{photometric} \gets D_{i}, D_{i+1}, I_i, I_{i+1}, P, K$
        \State $error_{geometric} \gets D_{i}, D_{i+1}, P, K$
        \State $current\_error=error_{photometric} + \alpha \times error_{geometric}$, $\alpha=0.5$
        \If{$current\_error<lowest\_error$ or $n=0$}
            \State $lowest\_error \gets current\_error$
            \State $best\_params \gets current\_params$
        \EndIf
        \State Update $current\_params$ with back propagation
    \EndFor
    \State Update DepthNet and PoseNet with $best\_params$, output results inferred with updated models
\EndFor
\end{algorithmic}
\end{algorithm}

\begin{figure*}[t]
    \centering
    \includegraphics[width=1\linewidth]{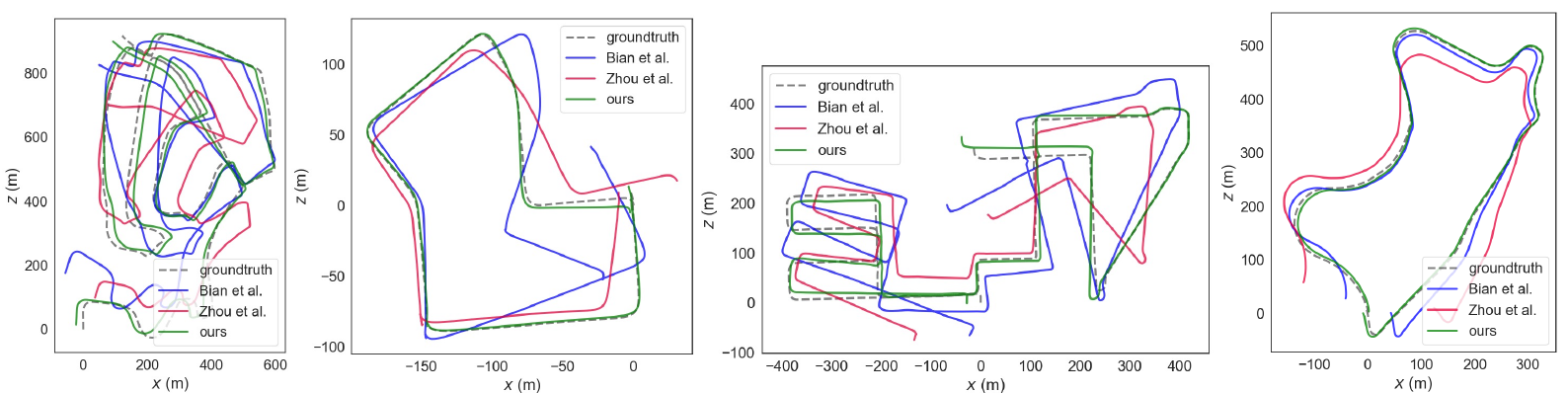}
    \caption{Qualitative results on the KITTI Odometry 02 07 08 and 09.
    Although the three compared methods adopt similar self-supervision and network designs, our method predicts a global trajectory that aligns more closely with the ground-truth, without experiencing trajectory drift over longer predictions.}
    \label{fig:traj}
\end{figure*}

\begin{table*}[htbp]
    \centering
    \scriptsize
    \caption{Quantitative comparison (Absolute Trajectory Error/Translation error/Rotation error) of methods on the KITTI Odometry benchmark. Both our two ways of using ORB features prove effective and outperform other compared methods. Even without SOA, our method surpasses the baseline method SC-SfMLearner by a significant margin.}
    \renewcommand{\arraystretch}{0.3}
    \setlength{\tabcolsep}{5pt} % Adjust cell spacing
    \label{tab:vo}
    \begin{tabular}{cc ccccccccccc}
        \toprule
        \textbf{Method} & \textbf{Metric} & \textbf{00} & \textbf{01} & \textbf{02} & \textbf{03} & \textbf{04} & \textbf{05} & \textbf{06} & \textbf{07} & \textbf{08} & \textbf{09} & \textbf{10} \\
        \midrule
        \multirow{3}{*}{\makecell[c]{SfMLearner~\cite{zhou2017unsupervised}}} & ATE         & 138.02 & 53.71 & 113.91 & 15.24 & 2.65 & 50.99 & 32.07 & 17.83 & 71.67 & 67.48 & 19.77 \\
                                                & Trans. err.  & 27.59 & 16.14 & 19.45 & 13.50 & 3.63 & 10.92 & 11.01 & 12.87 & 13.69  & 16.28 & 9.97 \\
                                                & Rot. err.    & 6.73 & 2.21 & 4.27 & 7.43 & 2.98 & 4.18 & 3.78 & 6.84 & 4.18 & 4.23 & 5.02 \\
        \midrule
        \multirow{3}{*}{\makecell[c]{SC-SfMLearner~\cite{jia2021scsfm}}} & ATE       & 108.77 & 549.41 & 105.31 & 10.75 & 2.50 & 49.82 & 13.22 & 28.37 & 60.18 & 31.25 & 14.11 \\
                                                  & Trans. err. & 12.39 & 165.86 & 8.44 & 10.11 & 4.45 & 7.85 & 3.78 & 13.06 & 11.31 & 8.98 & 10.45 \\
                                                  & Rot. err.   & 3.73 & 3.04 & 3.11 & 5.42 & 4.12 & 3.30 & 1.84 & 8.75 & 3.56 & 3.07 & 3.73 \\
        \midrule
        \multirow{3}{*}{\makecell[c]{Monodepth2~\cite{godard2019digging}}} & ATE       & 133.58 & 32.99 & 135.47 & 12.72 & 1.56 & 59.40 & 11.13 & 18.01 & 93.74 & 57.13 & 18.34 \\
                                                  & Trans. err. & 17.65 & 8.78 & 11.97 & 11.88 & 4.06 & 8.88 & 4.86 & 8.15 & 11.58 & 12.42 & 10.66 \\
                                                  & Rot. err.   & 3.79 & 1.16 & 2.16 & 5.99 & 1.49 & 3.95 & 1.56 & 5.14 & 4.00 & 2.72 & 4.53 \\
        \midrule
        \multirow{3}{*}{\makecell[c]{TartanVO~\cite{pmlr-v155-tartan}}} & ATE               & 63.45 & 70.79 & 67.64 & 7.72 & 2.90 & 54.07 & 24.62 & 19.56 & 59.36 & 31.97 & 25.08 \\
                                           & Trans. err.        & 9.94 & 18.87 & 9.87 & 7.16 & 8.08 & 9.40 & 9.50 & 9.78 & 11.43 & 10.03 & 13.41 \\
                                           & Rot. err.          & 3.59 & 1.93 & 3.37 & 2.73 & 4.47 & 3.24 & 2.51 & 4.96 & 3.17 & 3.18 & 3.21 \\

        \midrule
        \multirow{3}{*}{\makecell[c]{Ours (concatenate)}} & ATE               & 45.85  & 85.99 & 40.63 & 7.07 & 1.92 & 19.48 & 7.95 & 19.08 & 23.25 & 7.88 & 12.64 \\
                                           & Trans. err.        & 5.52 & 17.24 & 4.27 & 8.57 & 2.58 & 3.50 & 3.20 & 8.10 & 5.64 & 5.19 & 7.10 \\
                                           & Rot. err.          & 2.34 & 2.00 & 1.77 & 4.53 & 2.24 & 2.53 & 1.26 & 4.65 & 2.56 & 2.21 & 3.40 \\
        \cmidrule(lr){3-13}                    
        \multirow{3}{*}{\makecell[c]{Ours (attention)}} & ATE               & 33.45 & 34.24 & \textbf{29.73} & 5.06 & \textbf{1.84} & 19.53 & 9.26 & 14.12 & 20.66 & 9.26 & 12.39 \\
                                           & Trans. err.        & 5.47 & 10.11 & \textbf{4.12} & 7.26 & 2.57 & 4.47 & 4.37 & 7.45 & 6.75 & 5.61 & 8.00 \\
                                           & Rot. err.          & 2.27 & 1.18 & \textbf{1.65} & 3.47 & 1.80 & 2.73 & 1.65 & 4.25 & 2.64 & 2.16 & 3.44 \\
        \cmidrule(lr){3-13}                    
                                           
        \multirow{3}{*}{\makecell[c]{Ours (concatenate, with SOA)}} & ATE               & \textbf{28.62}  & 22.63 & 38.27 & \textbf{3.38} & 1.95 & \textbf{15.65} & 8.01 & 8.30 & 17.84 & 7.88 & \textbf{9.60} \\
                                           & Trans. err.        & 5.43 & 5.01 & 4.71 & \textbf{5.26} & \textbf{2.35} & \textbf{3.27} & \textbf{2.91} & 3.83 & 5.73 & 4.11 & \textbf{5.74} \\
                                           & Rot. err.          & \textbf{1.70} & \textbf{0.66} & 1.76 & \textbf{2.54} & \textbf{1.34} & \textbf{1.79} & \textbf{1.24} & 2.45 & 2.05 & 1.69 & 2.50 \\
        \cmidrule(lr){3-13}                    

        \multirow{3}{*}{\makecell[c]{Ours (attention, with SOA)}} & ATE               & 34.63  & \textbf{17.09} & 43.18 & 3.74 & 2.27 & 15.77 & \textbf{7.32} & \textbf{3.52} & \textbf{17.70} & \textbf{7.15} & 10.50 \\
                                           & Trans. err.        & \textbf{5.23} & \textbf{4.47} & 4.66 & 5.76 & 3.29 & 4.05 & 2.97 & \textbf{2.39} & \textbf{4.88} & \textbf{3.85} & 5.98 \\
                                           & Rot. err.          & 1.90 & 0.71 & 1.80 & 2.85 & 1.95 & 2.02 & 1.39 & \textbf{2.24} & \textbf{1.85} & \textbf{1.67} & \textbf{2.44} \\
        % \cmidrule(lr){3-13}                    

        \bottomrule
        \addlinespace[1pt]
        \multicolumn{13}{l}{The best performance for ATE, Trans. Err. (\%) and Rot. Err. (\degree/100m) is highlighted in \textbf{bold}.}
    \end{tabular}
    \vspace{-10pt}
\end{table*}

\section{EXPERIMENTS}
\subsection{Implementation Details}
We use SC-SfMLearner~\cite{jia2021scsfm} as the baseline method. 
The DepthNet is a U-Net~\cite{ronneberger2015u} structure with a ResNet-50~\cite{he2016deep} encoder. 
Our PoseNet embedded with multi-head cross attention layers use two ResNet-18 encoders for RGB and ORB feature extraction respectively. For a fair comparison, all the methods we compare use a ResNet-50 depth encoder, and are trained on the KITTI-Raw Eigen split~\cite{eigen2014depth} with image size set to $832 \times 256$. 
Our experimental setup is based on Python 3.7, PyTorch 1.11.0, and CUDA 11.8, and all experiments were conducted using an NVIDIA A100 GPU.
The networks were trained for 200,000 iterations and the learning rate was set to $1 \times 10^{-4}$ during both training and online adaptation.

\subsection{Data Preparation}
For ego-motion estimation, we use the KITTI-Raw dataset~\cite{Geiger2012CVPR} for training, following the Eigen split~\cite{eigen2014depth}. 
PoseNet and DepthNet are trained jointly on 42,440 images from 68 training scenes, leaving the remaining 10 scenes for validation, as done in previous work~\cite{jia2021scsfm,zhou2017unsupervised}. 
For testing, we select the KITTI Odometry dataset, which includes 11 sequences with ground truth camera poses. As noted by Yang~\emph{et al.}~\cite{yang2018deep}, sequences 01, 02, 06, 08, 09, and 10 are included in the training set of the Eigen split. 
Therefore, while the other sequences serve as strictly defined test sets, the sequences present in the training set also offer valuable insights into the performance of self-supervised models. 
Consequently, we include all the 11 sequences in our evaluation.

To verify the contribution of online adaptation to the generalization ability of the method, we selected the Virtual KITTI 2 (vKITTI)~\cite{cabon2020vkitti2} dataset for experiments across different domains. 
The advantage of the vKITTI dataset lies in its simulation of five different weather conditions along the same route, effectively providing five distinct domains with significant variability. 
Since the KITTI-Raw dataset is primarily collected under sunny conditions, this comparison highlights the method's generalization performance.

% \subsection{Evaluation Metric}

\subsection{Camera Ego-motion Estimation}
In TABLE~\ref{tab:vo}, we present a comparison of our method with other self-supervised VO approaches, along with an ablation study of our method. 
In addition to Absolute Trajectory Error (ATE), we compute Translation error (Trans. err.) and Rotation error (Rot. err.) for subsequences of length (100, ..., 800 meters) following KITTI~\cite{Geiger2012CVPR} official metrics. 
On the 11 testing sequence, our method demonstrates more accurate pose estimation. 
Notably, the baseline method SC-SfMLearner performs poorly on the 00-02 trajectories due to error accumulation in longer trajectories and domain gaps caused by factors such as higher vehicle speeds on highway conditions (sequence 01), whereas our method predicts these trajectories much more accurately. 
We tested our method without SOA and with two approaches for utilizing ORB features: (1) concatenating ORB features as explained in~\ref{ORB Aug}, and (2) feature fusion using cross-attention. 
If not otherwise specified, all other reported results are based on the second approach. 
Results of qualitative comparison of our method, Zhou~\emph{et al.}\cite{zhou2017unsupervised} and Bian~\emph{et al.}\cite{jia2021scsfm} are shown in Fig.~\ref{fig:traj}.

For online adaptation, we input three frames at a time and perform $k=2$ rounds of iteration on these frames. 
To further validate the effectiveness of our SOA, we conduct an ablation study. 
We test the results on sequences 07 and 09 under the following conditions: (1) varying numbers of iterations, (2) without the selective strategy (optimize the parameters without selecting the best ones based on self-supervised losses), and (3) different numbers of frames input at each step. 
TABLE~\ref{tab:online} shows the results of online adaptation at 4 different configurations. 
We observe that the proposed selective adaptation strategy effectively improves the pose estimation precision. 
In addition, increasing the number of iterations or the snippet sequence length does not necessarily improve prediction accuracy. 
Therefore, in our other experiments, we adopt the more efficient configuration of using two iterations with a snippet length of three.

\begin{table}[htbp]
    \centering
    \scriptsize
    \caption{Ablation study results on sequences 07 and 09, where ``Iteration'' refers to the number of iterations the network performs with each input during online adaptation, ``Selective'' indicates whether the parameters are updated based on the minimum self-supervised error, and ``Frames'' denotes the number of frames in each snippet of input.}
    \renewcommand{\arraystretch}{0.3}
    \setlength{\tabcolsep}{4pt} % Adjust cell spacing
    \label{tab:online}
    \begin{tabular}{cccc ccc}
        \toprule
        \textbf{Iteration} & \textbf{Selective} & \textbf{Frames} & \textbf{Metric} & \textbf{07} & \textbf{09} \\
        \midrule
        \multirow{3}{*}{\makecell[c]{2}} & \multirow{3}{*}{Yes} & \multirow{3}{*}{3} & ATE       & \textbf{3.52} & 7.15 \\
        &  &  & Trans. err. & \textbf{2.39} & \textbf{3.85} \\
        &  &  & Rot. err.   & 2.24 & 1.67 \\
        \midrule
        \multirow{3}{*}{\makecell[c]{2}} & \multirow{3}{*}{No} & \multirow{3}{*}{3} & ATE       & 5.56 & 16.01 \\
        &  &  & Trans. err. & 4.43 & 4.96 \\
        &  &  & Rot. err.   & 3.37 & 1.69 \\
        \midrule
        \multirow{3}{*}{\makecell[c]{2}} & \multirow{3}{*}{Yes} & \multirow{3}{*}{5} & ATE       & 5.29 & \textbf{6.81} \\
        &  &  & Trans. err. & 3.36 & 4.27 \\
        &  &  & Rot. err.   & 2.59 & 1.90 \\
        \midrule
        \multirow{3}{*}{\makecell[c]{3}} & \multirow{3}{*}{Yes} & \multirow{3}{*}{3} & ATE       & \textbf{3.52} & 8.75 \\
        &  &  & Trans. err. & 2.41 & 4.49 \\
        &  &  & Rot. err.   & \textbf{2.21} & \textbf{1.62} \\
        \bottomrule
        % \multicolumn{7}{l}{The best performance for ATE (m), Trans. Err. (\%) and Rot. Err. (\degree/100m) is}
        % \multicolumn{7}{l}{highlighted in \textbf{bold}.}
    \end{tabular}
    \vspace{-5pt}
\end{table}

\subsection{Attention Weights Visualization}
Our PoseNet with cross-attention layers takes two images concatenated along the channel axis as input. 
After ORB augmentation, features are extracted by the ORB encoder, and RGB features are extracted by the RGB encoder. 
We apply 8-head cross-attention between the deepest features from both encoders, and visualize the average attention weights across all heads, as shown in Fig.~\ref{fig:cross_vis}. 
The average attention weights are first transposed and averaged across the last dimension, reducing the 2D matrix from its original size of $208 \times 208$ to a single dimension of 208. 
This dimension is reshaped into $8 \times 26$ and resized to align the input image.

\begin{figure}[t]
    \centering
    \includegraphics[width=1\linewidth]{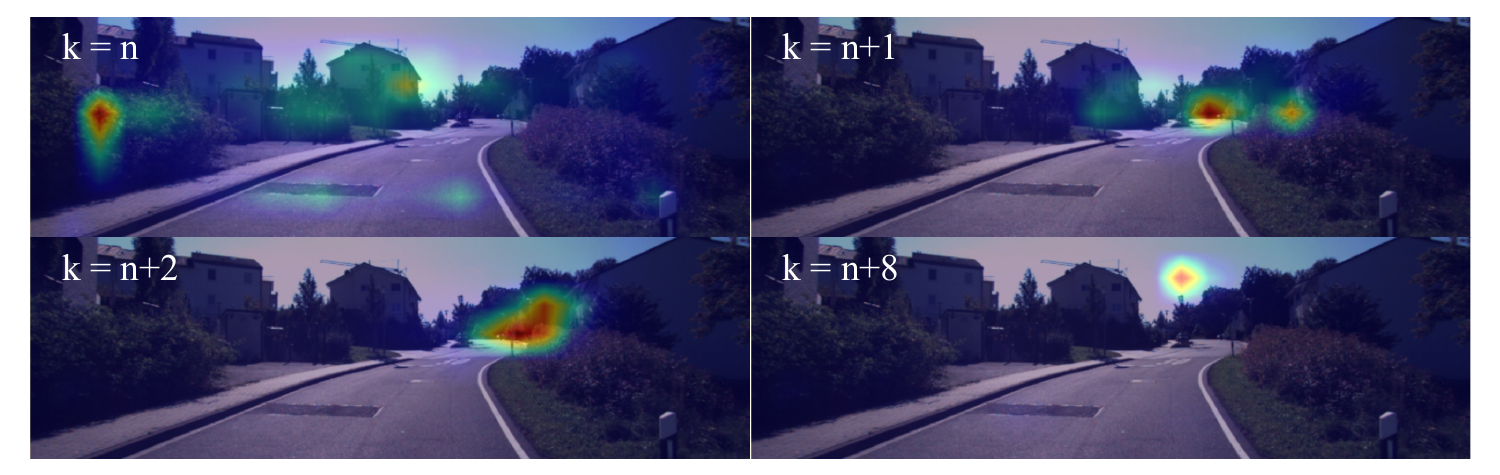}
    \caption{We examine the impact of the difference between two frames on the attention weights. 
    Setting the $n_{th}$ frame as the previous frame and the $k_{th}$ frame as the subsequent one, we observe that only with a moderate distance between n and k (i.e., moderate motion between frames), the highlighted areas point towards the distance of the road.}
    \label{fig:cross_interval}
\end{figure}

In addition to examining the patterns of attention weights between every two consecutive frames in a sequence, we also investigate how the extent of the frame interval affects the attention weights.
Fig.~\ref{fig:cross_interval} shows how the attention weights vary when setting the $n_{th}$ and $k_{th}$ frames as inputs. 
We find that when the interval between two frames is small, such as during complete stillness, the highlighted regions in the attention weights are often distributed across the entire image. 
As the interval increases, the attention focuses more on the end of the road.
This pattern aligns with the intuition from Fig.~\ref{fig:cross_vis} that it is difficult to determine if a car is turning left or right when stationary. 
In addition, as the frame interval increases too much, the highlighted regions also shift.

\subsection{KITTI to vKITTI}
We select three weather conditions ($Clone$, $Fog$, and $Morning$) from vKITTI Scene 01 and 20. The clone condition replicates the weather of KITTI.
We also find out the corresponding real sequences from KITTI for comparison. 
The pose estimation results are shown in TABLE~\ref{tab:vkitti}. 
The models perform better on KITTI because its data distribution is similar to that of our training set. 
However, even the conditions in vKITTI $Clone$ can have a significant impact on the accuracy of pose estimation. 
Our proposed SOA effectively overcomes this domain gap and performs equally well under three distinct weather conditions.

\begin{table}[t]
    \centering
    \scriptsize
    \caption{Generalizability test on scene 01 and 20 of the vKITTI dataset. Our full method achieves consistently good results across different domains.}
    \renewcommand{\arraystretch}{0.3}
    \setlength{\tabcolsep}{2pt} % Adjust cell spacing
    \label{tab:vkitti}
    \begin{tabular}{cc cccc cccc}
        \toprule
        & & \multicolumn{4}{c}{\textbf{Scene 01}} & \multicolumn{4}{c}{\textbf{Scene 20}} \\
        \cmidrule(lr){3-6} \cmidrule(lr){7-10}
        \textbf{Method} & \textbf{Metric} & \textbf{K} & \textbf{C} & \textbf{F} & \textbf{M} & \textbf{K} & \textbf{C} & \textbf{F} & \textbf{M} \\
        \midrule
        \multirow{3}{*}{SfMLearner} & ATE        & 35.58  & 35.77  & 24.32  & 35.15         & 16.62 & 16.40 & 17.64 & 15.33 \\
                                    & Trans. err. & 53.08  & 54.16  & 36.89  & 52.22  & 11.27 & 11.75 & 13.95 & 9.68 \\
                                    & Rot. err.   & 34.52  & 35.99 & 22.82 &  34.87  & 3.51 & 4.58 & 5.26 & 3.96 \\
        \midrule
        \multirow{3}{*}{SC-SfMLearner} & ATE        & 4.62 & 4.58 & 7.59 & 5.58        & 13.39 & 13.51 & 51.74 & 12.72 \\
                                      & Trans. err. & 4.80 & 5.27 & 9.63 & 6.38 & 11.80 & 9.20 & 33.52 & 9.18 \\
                                      & Rot. err.   & 4.53 & 4.09 & 6.46 & 5.13   & 3.91 & 1.95 & 7.97 & 1.66 \\
        \midrule
        \multirow{3}{*}{Monodepth2} & ATE        & 2.93 & 2.48 & 3.92 & 3.54        & 7.87 & \textbf{8.07} & 12.32 & 14.00 \\
                                      & Trans. err. & 4.46 & 3.02 & 4.62 & 3.15 & 7.79 & \textbf{6.79} & 11.11 & 13.10 \\
                                      & Rot. err.   & 3.20 & 2.76 & 5.89 & 3.46   & 2.43 & 2.64 & 3.92 & 4.41 \\
        \midrule
        \multirow{3}{*}{TartanVO} & ATE        & 6.67 & 6.15 & 6.59 & 6.36        & 23.65 & 16.25 & 22.37 & 17.56 \\
                                      & Trans. err. & 10.12 & 8.83 & 8.28 & 9.44 & 16.42 & 9.84 & 18.04 & 13.07 \\
                                      & Rot. err.   & 4.49 & 5.37 & 5.54 & 4.84   & 5.72 & 4.29 & 3.81 & 3.46 \\
        \midrule
        \multirow{3}{*}{Ours w/o SOA} & ATE        & 2.26 & 2.58 & 5.58 & 6.47         & 7.91 & 12.04 & 44.13 & 15.63 \\
                              & Trans. err. & 2.97 & 4.85 & 8.17 & 10.21 & 6.35 & 9.13 & 37.15 & 9.68 \\
                              & Rot. err.   & 3.11 & 3.20 & 2.76 & \textbf{3.05} & 2.48 & 3.59 & 13.39 & 3.64 \\
        \midrule
        \multirow{3}{*}{Ours} & ATE        & \textbf{1.38} & \textbf{1.96} & \textbf{1.18} & \textbf{1.48}       & \textbf{5.78} & 8.98 & \textbf{10.61} & \textbf{10.63} \\
                                   & Trans. err. & \textbf{2.16} & \textbf{2.99} &  \textbf{2.42} & \textbf{2.83} & \textbf{6.33} & 7.47 & \textbf{6.63} & \textbf{8.59} \\
                                   & Rot. err.   & \textbf{3.00} & \textbf{2.63} & \textbf{2.67} & 3.11 & \textbf{2.22} & \textbf{2.60} & \textbf{2.64} & \textbf{2.82} \\
        \bottomrule
        \addlinespace[1pt]
        \multicolumn{10}{l}{Dataset: \textbf{K}=KITTI, \textbf{C}=Clone-vKITTI, \textbf{F}=Fog-vKITTI, and \textbf{M}=Morning-vKITTI.}
    \end{tabular}
    \vspace{-10pt}
\end{table}

\section{CONCLUSION}
This paper proposes an ORB-augmented VO with SOA. ORB features guide the attention of the network to key regions, and more stable information from the raw data are used to estimate ego motion. Consequently, our model demonstrates superior pose accuracy across all trajectories in the KITTI odometry dataset.
Further, cross-attention module illustrates how ORB features guide the extraction of RGB features, providing a level of interpretability. We observe that regions with higher attention weights correspond to the vehicle's turning direction. Moreover, our ORB-SfMLearner works well under different domains due to optimal parameters adaptation. Overall, the integration of ORB augmentation has improved the accuracy and explainability of our model's pose estimation, while the SOA has further enhanced its generalization capabilities.

\bibliographystyle{IEEEtran}

\bibliography{references}

\end{document}